\documentclass[preprint,12pt]{elsarticle}

\usepackage{amssymb}
\usepackage{amsmath}
\usepackage{algorithmic}
\usepackage{graphicx}
\usepackage{textcomp}
\usepackage{algorithm}
\usepackage{amsfonts}
\usepackage{cleveref}
\usepackage{multirow} 

\journal{Expert Systems With Applications}

\begin{document}

\begin{frontmatter}



\title{SimDiff: Depth Pruning via Similarity and  Difference}


\author[label1]{Yuli Chen, Shuhao Zhang, Fanshen Meng, Bo Cheng\corref{cor1}} 
\author[label2] {Jiale Han}

\author[label3,label4]{Qiang Tong, Xiulei Liu}

\cortext[cor1]{Corresponding author: Bo Cheng (email: chengbo@bupt.edu.cn)}

\affiliation[label1]{organization={State Key Laboratory of Networking and Switching Technology,  Beijing University of Posts and Telecommunications},
            city={Beijing},
            postcode={100876}, 
            country={China}}
\affiliation[label2]{organization={Hong Kong University of Science and Technology},
            city={Hongkong},
            postcode={999077}, 
            country={China}}
            
\affiliation[label3]{organization={Laboratory of Data Science and Information Studies,  Beijing Information Science and Technology University},
            city={Beijing},
            postcode={100101}, 
            country={China}}
\affiliation[label4]{organization={Beijing Advanced Innovation Center for Materials Genome Engineering, Beijing Information Science and Technology University},
            city={Beijing},
            postcode={100101}, 
            country={China}}

\begin{abstract}
Depth pruning improves the deployment efficiency of large language models (LLMs) by identifying and removing redundant layers. A widely accepted standard for this identification process is to measure the similarity between layers using cosine distance. However, we find that methods relying solely on this one-dimensional heuristic can exhibit unpredictable performance and even catastrophic collapse across different architectures. To address this issue, we propose SimDiff, a novel layer importance criterion that jointly evaluates layers from two orthogonal perspectives: representational similarity and transformation difference. The difference is quantified using two distinct metrics: MSSD, which is sensitive to outliers and identifies layers that make decisive corrections, and MASD, which robustly measures a layer's average contribution. Extensive experiments on multiple models ranging from 0.5B to 13B parameters demonstrate that SimDiff significantly outperforms state-of-the-art baselines across various pruning ratios. Notably, our method retains over 91\% of LLaMA2-7B's performance at a 25\% pruning ratio and achieves up to a 1.49x inference speedup when pruning 12 layers on LLaMA3.1-8B. We also show that pruned models can be effectively recovered with minimal fine-tuning.
\end{abstract}

\begin{graphicalabstract}
\includegraphics[width=\columnwidth]{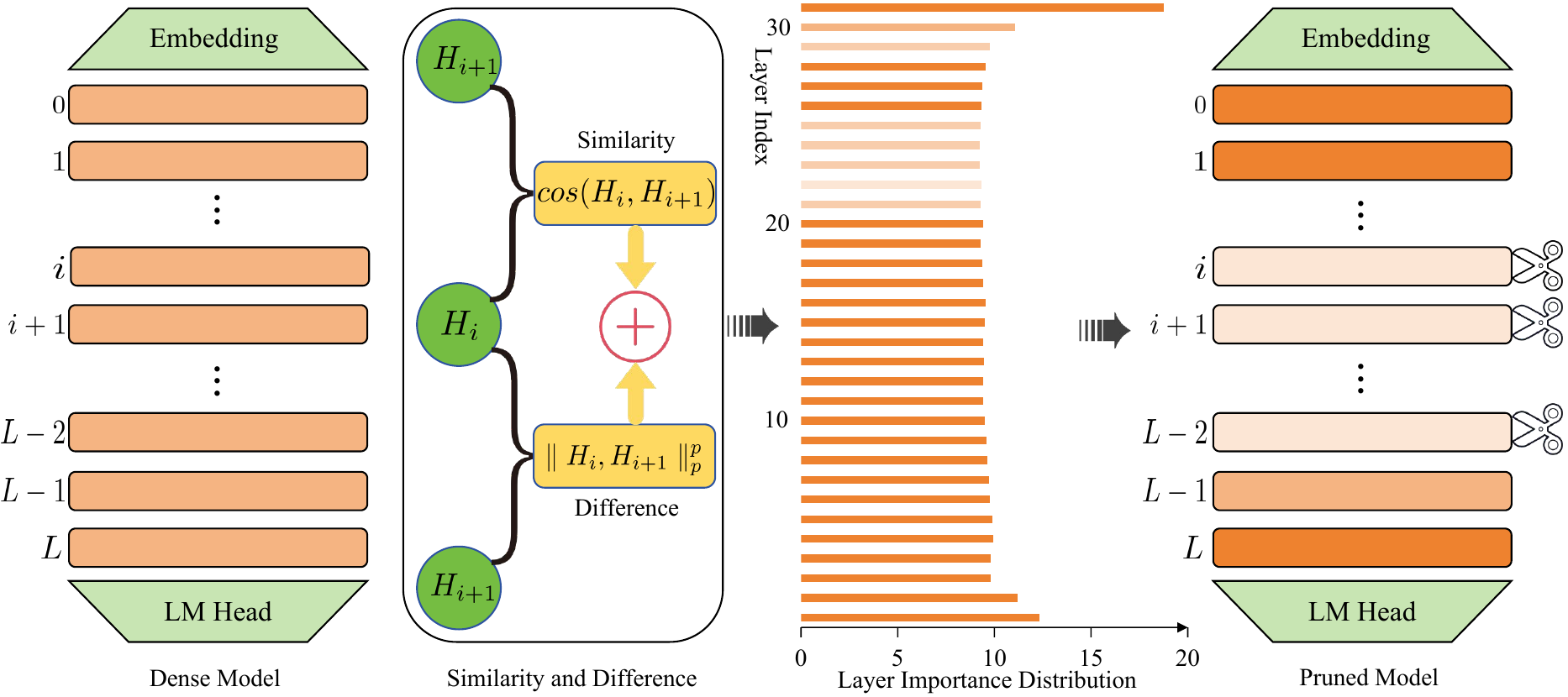}
\end{graphicalabstract}


\begin{highlights}
\item  SimDiff prunes LLMs by jointly measuring inter-layer similarity and difference.
\item Two metrics, MSSD and MASD, capture distinct and complementary layer behaviors.
\item SimDiff preserves >91\% performance at 25\% pruning and speeds up inference 1.49×.
\end{highlights}


\end{frontmatter}



\section{Introduction}
\label{sec:introduction}

The rapid scaling of large language models (LLMs) and their associated heavy resource demands hinder deployment in resource-constrained settings, which has spurred the development of effective compression techniques \cite{jaiswal2023the}. Among these, pruning \cite{ma2023llmpruner, men2024shortgpt} is a leading approach. It removes redundant components to shrink model size and accelerate inference while maintaining accuracy. Pruning strategies are often categorized as either unstructured \cite{pmlr-v202-frantar23a} or structured \cite{YangC024, SongOKKKK24}. The latter, which removes entire components for hardware-friendly acceleration, can be further divided into two main approaches: Width pruning \cite{ma2023llmpruner, ashkboos2024slicegpt} and depth pruning \cite{men2024shortgpt, YangC024}. Width pruning removes substructures such as attention heads or MLP units within each layer. While width pruning addresses the computational cost within each layer, it preserves the model's depth. Depth pruning, conversely, removes entire layers to shorten the sequential computation path directly. This fundamental difference makes depth pruning a more effective strategy for latency-sensitive applications in real-time or resource-limited scenarios.

\begin{figure}[t!]
\centering
\includegraphics[width=0.8\columnwidth]{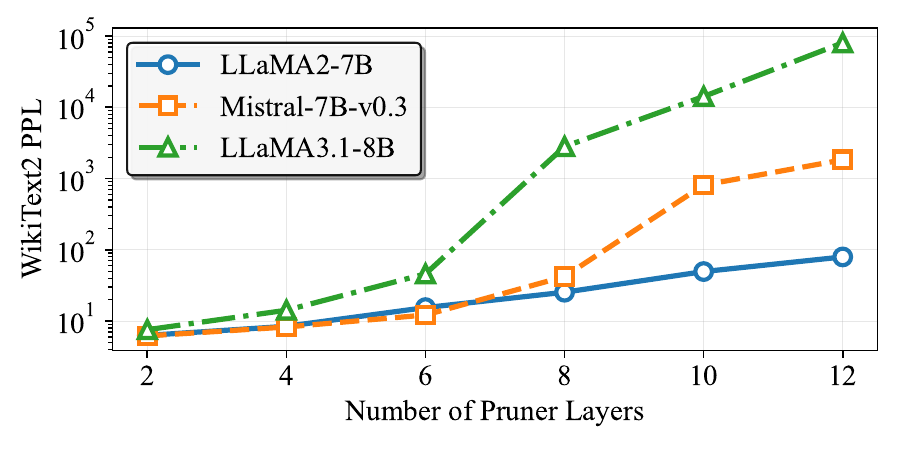} 
\caption{WikiText2 perplexity (PPL) for three models across varying numbers of pruned layers using ShortGPT. The y‑axis uses a logarithmic scale, and lower PPL indicates better language modeling performance.}
\label{motivation}
\end{figure}

\begin{figure}[t!]
\centering
\includegraphics[width=\textwidth]{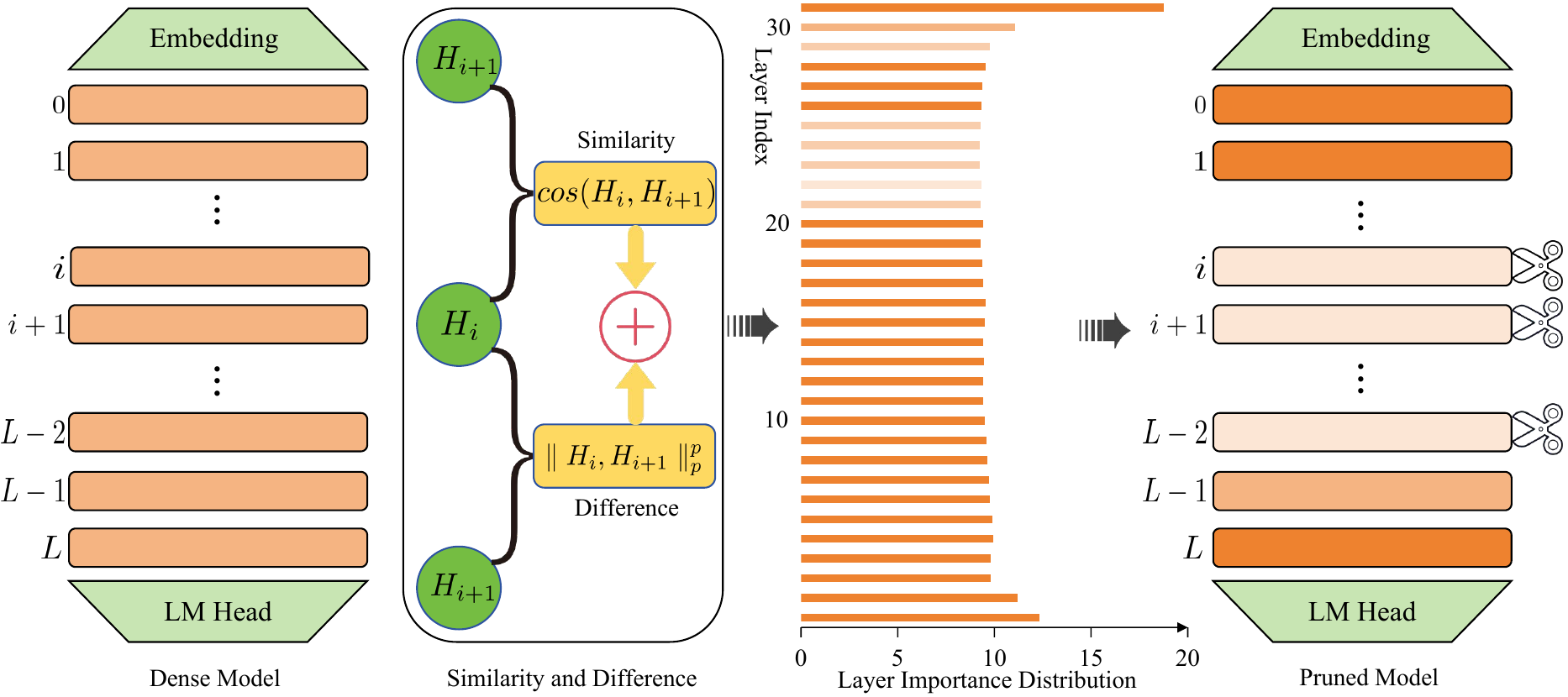} 
\caption{Illustration of SimDiff. Layer importance is computed based on both similarity and difference metrics. The layers are then ranked accordingly, and those with lower importance scores are pruned. Lighter colors indicate lower importance.}
\label{overall}
\end{figure}

Recent advances \cite{YangC024, men2024shortgpt} in depth pruning introduce various strategies for assessing layer importance, with ShortGPT \cite{men2024shortgpt} being a notable example due to its simple use of cosine similarity to capture representational redundancy. However, this simplicity conceals a critical lack of robustness, as illustrated in  Figure \ref{motivation}. The figure shows the performance of three models when pruned with ShortGPT, revealing a stark divergence in outcomes: LLaMA2-7B \cite{touvron2023llama} maintains relative stability, while both Mistral-7B-v0.3 \cite{thakkar2023comprehensive}and LLaMA3.1-8B  undergo an abrupt performance collapse after a critical number of layers are removed. This architectural dependency highlights the core deficiency of using a simplistic metric like cosine similarity. The unpredictable nature of its effectiveness suggests fundamental limitations, and we hypothesize that these may stem from two issues: first, its tendency to focus solely on directionality while potentially ignoring the magnitude of a layer's transformation, and second, its possible sensitivity to alignment noise, which could lead to instability.

To address these limitations, we propose a novel depth pruning method, SimDiff, which jointly considers the similarity and difference between adjacent hidden states. The pipeline of SimDiff is illustrated in Figure \ref{overall}. Specifically, SimDiff uses cosine distance to measure similarity and introduces two distinct metrics to evaluate differences. MASD assumes that smaller changes between adjacent layers indicate lower importance and thus measures the average contribution of each layer. MSSD amplifies larger deviations between layers to identify critical ones, making it more sensitive to outliers. This principle aligns with recent studies such as OWL \cite{yin2024outlier} and DLP \cite{chen2025dlp}, which suggest that layers with more outliers are more important. By integrating these two orthogonal perspectives, SimDiff offers a more comprehensive and robust criterion for identifying redundant layers, thereby improving the reliability of pruning and the performance on downstream tasks. To evaluate the effectiveness of SimDiff, we conduct comprehensive experimental evaluations across mainstream LLMs. Notably, SimDiff consistently outperforms the state-of-the-art LLM pruning techniques. Furthermore, when post-training is applied, only a small number of samples are needed to restore model performance rapidly.

Overall, the contributions of our work are as follows: 
\begin{itemize}
\item We propose SimDiff, a depth pruning method that jointly measures layer similarity and difference to form a unified and robust importance criterion.
\item We introduce two complementary difference metrics, MSSD and MASD, capturing both outlier-sensitive and stable transformation behaviors.
\item We design an adaptive weighting coefficient $\alpha$, optimized via ternary search, enabling architecture- and scale-aware pruning across models of different sizes.
\item Extensive experimental results across large language models from 0.5B to 13B parameters show that SimDiff consistently outperforms state-of-the-art pruning baselines. The pruned models not only achieve significant inference acceleration but also rapidly restore performance using only a small number of samples through lightweight LoRA fine-tuning.

\end{itemize}

\section{Related Work}


Model pruning is a widely used neural network compression technique that removes redundant parameters while maintaining model performance. Based on the granularity of pruning, existing methods are typically classified into unstructured \cite{pmlr-v202-frantar23a} and structured pruning \cite{YangC024, SongOKKKK24, men2024shortgpt, kim2024shortened}. Unstructured pruning removes individual weights according to importance metrics. It achieves high sparsity. However, it introduces irregular patterns that do not align well with modern hardware acceleration. In contrast, structured pruning eliminates larger components such as neurons, channels, attention heads, or entire layers, making it more compatible with practical deployment. Structured pruning can be further divided into width pruning and depth pruning. Width pruning focuses on reducing intra-layer computational cost by pruning substructures such as attention heads, channels, or MLP units. For instance, LLM-Pruner \cite{ma2023llmpruner} analyzes inter-parameter dependencies to define the smallest removable groups and performs multi-dimensional structured pruning, significantly reducing inference latency while maintaining accuracy. SliceGPT \cite{ashkboos2024slicegpt} proposes instance-specific path pruning based on token importance, enabling dynamic sparsity without modifying the model depth. However, these approaches still retain the full model depth, which limits their effectiveness in extreme compression scenarios. To address this, depth pruning methods \cite{YangC024, men2024shortgpt, kim2024shortened} have been introduced to compress models by removing entire layers, resulting in substantial reductions in sequential computation and faster inference.

Depth pruning has gained increasing attention in the context of LLMs due to its ability to shorten computation paths directly. Several recent works have proposed various criteria for assessing layer importance. LaCo \cite{YangC024} merges adjacent layers based on linear similarity, preserving representational diversity while reducing depth. ShortGPT \cite{men2024shortgpt} employs cosine similarity between layer outputs to detect redundancy. Shortened-LLM \cite{kim2024shortened} explores task-specific depth reduction strategies to produce more compact models. SLEB \cite{SongOKKKK24} introduces a learning-based mechanism for evaluating each block’s contribution to downstream tasks. Concurrently, EntroDrop \cite{yang2025entropy} evaluates the entropy of hidden states to quantify information contribution and prunes layers with low information value.

\section{Methodology}

In this section, we introduce SimDiff, a depth pruning method that first formulates the pruning objective, then proposes a unified importance metric based on similarity and difference, and finally details the algorithm for selecting and removing redundant layers.

\subsection{Problem Formulation}

Given an LLM $M = \{T_1, T_2, ..., T_L\}$ composed of $L$ layers, the objective of depth pruning is to remove a subset of layers $P \subset \{T_1, ..., T_L\}$ to obtain a shallower, pruned model $M'$ with $L' = L - |P|$ layers. This pruning process must satisfy two conditions: 1) maximizing the improvement in computational efficiency, and 2) minimizing the degradation of the model's original performance. To achieve this, we require an effective metric, $I(T_i)$, to evaluate the importance of the $i$-th layer $T_i$ and to prioritize the removal of layers with the lowest importance scores.

\subsection{Layer Importance}

LLMs are primarily built upon the Transformer architecture, which consists of a stack of decoder layers. Each of these layers adopts a residual structure. To analyze the $i$-th layer $T_i$, we denote its  hidden state as $H^{(i)}$. Here, $H \in \mathbb{R}^{B \times S \times D}$, where B, S, and D represent the batch size, sequence length, and hidden dimension, respectively. The transformation can be viewed as an update to the incoming hidden state:

\begin{equation}
H^{(i+1)} = H^{(i)} + \mathcal{F}_i(H^{(i)})
\label{eq:rev_1}
\end{equation}

Here, $\mathcal{F}_i(H^{(i)})$ is the non-linear transformation performed by the layer. This simple formula inspires us to evaluate layer importance from two perspectives:
\begin{enumerate}
    \item \textbf{Similarity}: Comparing how similar $H^{(i+1)}$ is to $H^{(i)}$. If they are highly similar, the layer is close to an identity transformation and its necessity may be low.
    \item \textbf{Difference}: Measuring the magnitude of the transformation $\mathcal{F}_i(H^{(i)})$. If the magnitude of this transformation is small, the layer is making only a minor modification to the signal, and its contribution may be limited.
\end{enumerate}
Our method quantifies both dimensions and fuses them into a unified importance score. 

\subsubsection{Similarity Metric: Identifying Representational Redundancy}

To measure whether the transformation performed by a layer is close to an identity transformation, we calculate the Cosine Similarity between its input and output hidden states. High cosine similarity indicates a minimal change in the direction of the input vectors. We define a "dissimilarity" term, $\mathcal{L}_{sim}^{(i)}$, which is inversely proportional to similarity:

\begin{equation}
\mathcal{L}_{sim}^{(i)} = 1 - \frac{1}{B \cdot S} \sum_{j=1}^{B \cdot S} \cos(H_{j}^{(i)}, H_{j}^{(i+1)})
\label{eq:rev_2}
\end{equation}
where $H_{j}^{(i)}$ and $H_{j}^{(i+1)}$ are the vectors for the $j$-th token from the flattened hidden states before and after the layer, respectively.

\subsubsection{Difference Metric: Quantifying Transformation Magnitude}

The difference metric aims to directly quantify the magnitude of the transformation function $\mathcal{F}_i(H^{(i)}) = H^{(i+1)} - H^{(i)}$. A larger transformation magnitude implies that the layer performs a more significant modification on the dataflow and thus may carry a more important function. We propose two alternative metrics to quantify this magnitude.

\paragraph{MSSD}
MSSD calculates the average of the squared magnitude of the differences between adjacent hidden states. It is defined as:

\begin{equation}
\mathcal{L}_{diff}^{(i)} = \frac{1}{B \cdot S} \sum_{j=1}^{B \cdot S} \| H_{j}^{(i+1)} - H_{j}^{(i)} \|_2^2
\label{eq:rev_3}
\end{equation}

MSSD squares the differences between adjacent hidden states, assigning disproportionately high weight to large transformations. This property makes the metric highly sensitive to statistical outliers. The emphasis on outliers forms the basis of our hypothesis. This idea also aligns with the principle behind recent methods \cite{yin2024outlier, chen2025dlp}, which identify layers with a higher concentration of outliers as more important. We hypothesize that the importance of a layer is reflected in its ability to produce decisive, high-magnitude changes. These changes often appear as outliers in the residual stream.

\paragraph{MASD}
MASD calculates the average of the absolute magnitude of the differences between adjacent hidden states. It is defined as:

\begin{equation}
\mathcal{L}_{diff}^{(i)} = \frac{1}{B \cdot S \cdot D} \sum_{j=1}^{B \cdot S} | H_{j}^{(i+1)} - H_{j}^{(i)} |
\label{eq:rev_4}
\end{equation}

Unlike MSSD, MASD measures the average magnitude of the differences between adjacent hidden states. It quantifies how much, on average, a layer modifies each hidden dimension, without being overly influenced by a few extreme values. In theory, MASD is statistically more robust and less sensitive to outliers. We hypothesize that the most important layers are those that consistently and steadily process the information stream, rather than those that perform occasional, drastic changes. MASD captures a layer's routine contribution to the model’s computation.

\subsubsection{Balanced Importance Score}

Similarity captures the directional alignment between adjacent hidden states, and difference measures the magnitude of change between them. To obtain a comprehensive layer importance score, we perform a weighted fusion of the similarity and difference dimensions. First, we uniformly apply the Sigmoid function to map the value into the $[0, 1]$ interval:

\begin{equation}
\mathcal{I}_{diff}(T_i) = \frac{1}{1 + e^{-\mathcal{L}_{diff}^{(i)}}}
\label{eq:rev_5}
\end{equation}

For the similarity metric $\mathcal{L}_{sim}^{(i)}$, its original value lies in the range $[0, 2]$. To get a score that is normalized to $[0, 1]$ and where a higher value means more dissimilar (and thus more important), we apply the following scaling:

\begin{equation}
\mathcal{I}_{sim}(T_i) = \frac{\mathcal{L}_{sim}^{(i)}}{2.0}
\label{eq:rev_6}
\end{equation}

Next, we introduce a hyperparameter $\alpha \in [0, 1]$ to balance the relative importance of the difference and similarity metrics. The final importance score $I(T_i)$ is defined as:

\begin{equation}
I(T_i) = \alpha \cdot \mathcal{I}_{diff}(T_i) + (1 - \alpha) \cdot \mathcal{I}_{sim}(T_i)
\label{eq:rev_7}
\end{equation}

In this composite metric, $\alpha$ acts as a weighting coefficient, allowing us to adjust the pruning criterion flexibly:
\begin{itemize}
    \item When $\alpha$ is close to 1, the pruning decision relies more heavily on the difference metric, prioritizing the preservation of layers that induce significant changes to the hidden states.
    \item When $\alpha$ is close to 0, the decision relies more on the similarity metric, prioritizing the preservation of layers whose outputs are most dissimilar from their inputs.
\end{itemize}

To efficiently determine the optimal hyperparameter $\alpha$, we employ a ternary search over the interval [0, 1], using perplexity as the optimization objective. The pseudocode of ternary search is provided in Algorithm \ref{alg:ternary_search_alpha}. Specifically, the search iteratively refines this interval. In each step, the algorithm prunes the model using two intermediate $\alpha$ values and evaluates their respective perplexity on a compact calibration set of 100 samples from the WikiText2 dataset. One-third of the interval corresponding to the higher perplexity is then discarded, allowing the search to rapidly converge on a near-optimal $\alpha$ that best balances our similarity and difference metrics.

\begin{algorithm}[tb]
\caption{Pseudocode of Ternary Search for Optimal Alpha}
\label{alg:ternary_search_alpha}
\textbf{Input}: Evaluation function $\text{EvaluatePPL}(\alpha)$ that returns a perplexity score\\
\textbf{Parameter}: Search precision $\epsilon$, max\_iterations $I_{max}$\\
\textbf{Output}: Optimal alpha value $\alpha^*$ that minimizes PPL
\begin{algorithmic}[1] 
\STATE Initialize search bounds: $left \leftarrow 0.0$, $right \leftarrow 1.0$
\STATE Initialize best found value: $\alpha^* \leftarrow \text{None}$, $PPL^* \leftarrow \infty$
\STATE Initialize iteration counter: $i \leftarrow 0$
\WHILE{$(right - left) > \epsilon$ \textbf{and} $i < I_{max}$}
    \STATE $i \leftarrow i + 1$
    \STATE \COMMENT{Calculate two midpoints to divide the interval into three sections}
    \STATE $m_1 \leftarrow left + (right - left) / 3$
    \STATE $m_2 \leftarrow right - (right - left) / 3$
    
    \STATE $PPL_1 \leftarrow \text{EvaluatePPL}(m_1)$
    \STATE $PPL_2 \leftarrow \text{EvaluatePPL}(m_2)$

    \STATE \COMMENT{Update the best-known value if a better one is found}
    \IF{$PPL_1 < PPL^*$}
        \STATE $PPL^* \leftarrow PPL_1$, $\alpha^* \leftarrow m_1$
    \ENDIF
    \IF{$PPL_2 < PPL^*$}
        \STATE $PPL^* \leftarrow PPL_2$, $\alpha^* \leftarrow m_2$
    \ENDIF

    \STATE \COMMENT{Discard the one-third of the interval with the higher PPL}
    \IF {$PPL_1 > PPL_2$}
        \STATE $left \leftarrow m_1$
    \ELSE
        \STATE $right \leftarrow m_2$
    \ENDIF
\ENDWHILE
\STATE \textbf{return} $\alpha^*$
\end{algorithmic}
\end{algorithm}

\subsection{Layer Removal}

Once the importance score, $I(T_i)$, has been calculated for all candidate layers $T_i$ in the model, we rank them based on this score to identify the most prunable layers. Specifically, a small value indicates lower importance. We rank all layers in ascending order based on their Importance Scores:

\begin{equation}
\text{R}(I) = \text{argsort}\left(\{I(T_i)\}_{i=1}^L\right)
\label{eq:rev_8}
\end{equation}
where $L$ is the total number of layers, and the `argsort` function returns the list of layer indices sorted by their importance scores in ascending order.

Finally, the $K$ layers with the smallest importance scores are selected to form the target pruning set $P$. Here, $K$ is the target number of layers to be pruned. The set of layers to be pruned, $P$, is thus defined as:

\begin{equation}
P = \{T_i \mid i \in \text{R}(I)[:K]\}
\label{eq:rev_9}
\end{equation}

Following this strategy, the layers within the set $P$ are identified as the least contributory and most redundant components of the model. Pruning them allows for maximal model depth compression with minimal negative impact on performance. The pseudocode of SimDiff is provided in Algorithm \ref{alg:SimDiff}.

\begin{algorithm}[tb]
\caption{Pseudocode of SimDiff}
\label{alg:SimDiff}
\begin{algorithmic}[1]
\STATE \textbf{Input}: Pretrained model $M = \{T_1, T_2, ..., T_L\}$, a calibration dataset $\mathcal{D}_{calib}$, number of layers to prune $K$
\STATE \textbf{Parameter}: Weighting coefficient $\alpha \in [0, 1]$
\STATE \textbf{Output}: Pruned model $M'$
\STATE Initialize an empty list for importance scores: $\mathcal{S} = []$.
\FOR{each layer $T_i$ in $M$ where $i=1, ..., L$}
    \STATE Obtain hidden states $H^{(i)}, H^{(i+1)}$ by running a forward pass on $\mathcal{D}_{calib}$.
    \STATE Calculate dissimilarity $\mathcal{L}_{sim}^{(i)}$ using Eq. \ref{eq:rev_2}.
    \STATE Calculate normalized similarity importance using Eq. \ref{eq:rev_6}.
    \STATE Calculate difference magnitude $\mathcal{L}_{diff}^{(i)}$ using either MSSD (Eq. \ref{eq:rev_3}) or MASD (Eq. \ref{eq:rev_4}).
    \STATE Calculate normalized importance $\mathcal{I}_{diff}(T_i) $ using Eq. \ref{eq:rev_5}.
    \STATE Compute the final importance score $I(T_i)$ using Eq. \ref{eq:rev_7}.
    \STATE Append the tuple $(I(T_i), i)$ to the list $\mathcal{S}$.
\ENDFOR
\STATE Get the ranked list of layer indices $\text{R}(I)$ using \cref{eq:rev_8}.
\STATE Select the set of layers to prune using \cref{eq:rev_9}.
\STATE Construct the pruned model $M'$ by removing the layers in set $P$ from $M$.
\STATE \textbf{return} $M'$
\end{algorithmic}
\end{algorithm}

\section{Experiments}

\subsection{Experimental Setup}

\subsubsection{Models and Datasets} 

To evaluate the effectiveness of our method, we conduct experiments on a diverse set of large language models, covering parameter scales from 0.5B to 13B. The evaluated models include LLaMA2-7B \cite{touvron2023llama}, LLaMA2-13B, Mistral-7B-v0.3 \cite{thakkar2023comprehensive}, LLaMA3.1-8B \cite{ling2024beemanc}, Qwen2.5-0.5B \cite{team2024qwen2}, Qwen2.5-1.5B, and Qwen2.5-3B. We evaluate two variants of our method, Ours (MSSD) and Ours (MASD), which use the MSSD and MASD metrics, respectively. Following prior work \cite{YangC024}, we prune 25\% of the layers from the LLaMA2-7B  model and evaluate the compressed models using the OpenCompass \cite{2023opencompass} framework across a wide range of natural language understanding (NLU) benchmarks, including CMNLI \cite{cmnli}, HeSW \cite{Zellers2019HellaSwagCA}, PIQA \cite{BiskZLGC20},  WSC \cite{levesque2012winograd}, CoQA \cite{ReddyCM19}, BoolQ \cite{clark-etal-2019-boolq}, Race-M and Race-H \cite{LaiXLYH17}, XSum \cite{HasanBIMLKRS21}, and C3 \cite{SunYYC20}. Additionally, we evaluate the language modeling capabilities and zero-shot performance. Specifically, we evaluate language modeling performance using the perplexity metric on the Wikitext2 dataset. For zero-shot evaluation, we assess accuracy on eight commonsense benchmarks from EleutherAI LM Harness \cite{eval-harness}, including BoolQ, RTE \cite{wang-etal-2018-glue}, HeSW, WinoG \cite{sakaguchi2021winogrande}, ARC-e and ARC-c \cite{boratko-etal-2018-systematic}, PIQA, MTQA \cite{amini2019mathqa}, and OBQA \cite{mihaylov-etal-2018-suit}.

\subsubsection{Baselines}  To evaluate the effectiveness of our work, we compare it against state-of-the-art structured pruning techniques. As our work is a depth pruning method, our primary comparison is with other leading depth pruning approaches, including LaCo \cite{YangC024}, ShortGPT \cite{men2024shortgpt}, Shortened-LLM \cite{kim2024shortened}, SLEB \cite{SongOKKKK24}, and EntroDrop \cite{yang2025entropy}. For a broader context, we also include representative width pruning methods like LLM-Pruner \cite{ma2023llmpruner} and SliceGPT \cite{ashkboos2024slicegpt}. To ensure a fair comparison, all baselines are evaluated without any post-training. 



\subsubsection{Implementation Details} 
We randoml sample 512 instances from the WikiText2 \cite{merity2016pointer} dataset as calibration data, with each instance having a sequence length of 2048. We further provide the post-training setup and recovery results in Appendix~\ref{section:lora}.

\begin{table}[t]
\centering

\caption{Comparison of pruning LLaMA2-7B for 25\% using different methods, without any healing. “*” indicates that we refer to the results in the LaCo paper. AVG is the average performance of the model on the datasets. Retained Performance (RP) represents the percentage of the original model’s performance retained by the pruning method. The best performance result is indicated in bold.}
\resizebox{\textwidth}{!}{
\begin{tabular}{ccccccccccccc}
\hline
\small{Method} & \small{CMNLI} & \small{HeSW} & \small{PIQA} & \small{WSC} & \small{CoQA} & \small{BoolQ} & \small{Race-M} & \small{Race-H} & \small{XSum} & \small{C3} & \small{AVG} & \small{RP} \\ \hline
\small{Dense} & 32.98 & 71.30 & 78.24 & 37.50 & 66.58 & 70.67 & 33.08 & 35.45 & 19.67 & 43.78 & 48.93 & 100.0 \\
\small{Dense*} & 32.99 & 71.26 & 77.91 & 50.00 & 64.62 & 71.62 & 35.71 & 34.19 & 19.40 & 43.56 & 50.13 & 100.0 \\
\small{LLMPruner*} & 34.33 & 56.46 & 71.22 & 36.54 & 42.51 & 55.20 & 22.56 & 22.35 & 11.51 & 25.64 & 37.83 & 75.46 \\
\small{SliceGPT*} & 31.70 & 50.27 & 66.21 & 36.54 & 41.36 & 38.32 & 21.07 & 21.66 & 4.89 & 39.78 & 35.18 & 70.18 \\
\small{LaCo*} & 34.43 & 55.69 & 69.80 & 40.38 & 45.70 & 64.07 & 22.61 & 23.61 & 15.64 & 39.67 & 41.16 & 82.11 \\
\small{ShortGPT} & 33.01 & 58.19 & 66.81 & 36.54 & 49.22 & 65.26 & 35.45 & 36.96 & 10.06 & 41.32 & 43.28 & 88.45 \\
\small{Ours (MSSD)} & 34.40 & 57.46 & 68.82 & 60.58 & 54.05 & 55.57 & 32.17 & 33.85 & 11.33 & 39.12 & \textbf{44.74} & \textbf{91.44} \\
\small{Ours (MASD)} & 33.80 & 56.14 & 67.14 & 42.31 & 49.14 & 64.62 & 34.40 & 33.70 & 10.96 & 41.75 & \textbf{43.40} & \textbf{88.70} \\ \hline
\end{tabular}
}
\label{table1}
\end{table}

\subsection{Main results}

\subsubsection{Natural Language Understanding Benchmark} 

To validate the effectiveness of the proposed method, we apply a 25\% pruning ratio to the LLaMA2-7B model using various pruning strategies and conduct a comparative evaluation on a suite of natural language understanding benchmarks within the OpenCompass framework. As shown in Table \ref{table1}, all methods are evaluated under the same setting without any healing strategies. The performance differences between Dense and Dense* across various datasets are relatively small, indicating that our experimental setup is largely consistent with that of LaCO \cite{YangC024}. To ensure fairness, we compare the results reported in the LaCO paper against Dense* when computing the Relative Performance (RP), while our experimental results are compared against Dense. The results indicate that our proposed methods exhibit a significant advantage in performance retention. Specifically, Ours (MSSD) achieves the highest average score of 44.74\% and the highest retained performance of 91.44\%, ranking first among all methods. Ours (MASD) also performs strongly, achieving 88.70\% RP, outperforming existing methods such as LaCo (82.11\%) and LLMPruner (75.46\%). In contrast, SliceGPT shows the weakest results, with only 70.18\% of the original performance retained. These findings robustly demonstrate that, under an identical 25\% pruning ratio, Ours (MSSD) not only preserves model performance more effectively but also generalizes well across diverse tasks. This highlights the superiority of our approaches in structured model compression for LLMs.

\subsubsection{Language Modeling and Zero-Shot Tasks} 

To further evaluate the generalizability of our method, we apply a 20\% pruning ratio to the Mistral-7B-v0.3 model and conduct both zero-shot evaluation and language modeling tasks. Notably, under this pruning ratio, MSSD and MASD produce identical pruned layer selections and are thus reported as a unified result. As shown in Table \ref{table2}, our work achieves the best overall performance, attaining an average score of 50.14\% and surpassing all other baselines. Regarding retained performance (RP), our approach preserves 81.16\% of the original model's capabilities, outperforming SliceGPT (75.96\%), ShortGPT (79.82\%), and substantially exceeding SLEB (55.45\%), which exhibits severe performance degradation. Additionally, on the WikiText2 language modeling task, our method yields a slightly higher perplexity than SliceGPT, which can be attributed to the fact that SliceGPT, as a width pruning method, preserves the full model depth and thus maintains stronger sequential dependency modeling. These results indicate that our method introduces less degradation in language modeling while preserving overall task performance, thereby demonstrating the effectiveness and robustness of the proposed pruning criterion.

\begin{table*}[t]
\centering
\caption{Result of pruning Mistral-7B-v0.3 for 20\% using different methods, without any healing. When MSSD and MASD produce identical pruned layer selections, we uniformly refer to them as Ours. AVG is the average performance of the model on the zero-shot datasets. Retained performance (RP) represents the percentage of the original model’s performance retained by the pruning method. The best performance result is indicated in bold.}
\resizebox{\textwidth}{!}{
\begin{tabular}{ccccccccccccc}
\hline
\small{Method} & \small{WikiText2}$\downarrow$  & \small{BoolQ} & \small{PIQA} & \small{HeSW} & \small{WinoG} & \small{ARC-e} & \small{ARC-c} & \small{OBQA} & \small{MTQA} & \small{AVG} & \small{RP} \\ \hline
\small{Dense} & 5.32  & 82.14 & 80.20 & 60.89 & 73.88 & 79.63 & 48.89 & 33.20 & 35.38 & 61.78 & 100.0 \\
\small{SliceGPT} & \textbf{6.46}  & 43.94 & 69.15 & 42.80 & 66.30 & 67.00 & 36.01 & 23.00 & 27.24 & 46.93 & 75.96 \\
\small{SLEB} & 123628  & 62.17 & 53.59 & 25.64 & 50.20 & 25.17 & 22.01 & 15.40 & 19.87 & 34.26  & 55.45 \\
\small{Shortened-LLM} & 67.41  & 69.72 & 67.41 & 39.74 & 65.59 & 52.40 & 33.62 & 21.40 & 24.59 & 46.81 & 75.77 \\
\small{ShortGPT} & 25.69  & 69.57 & 71.00 & 46.21 & 68.27 & 58.00 & 34.90 & 19.40 & 27.14 & 49.31 & 79.82 \\
\small{Ours} & 16.06  & 74.83 & 68.77 & 46.81 & 69.53 & 58.96 & 35.58 & 20.80 & 25.86 & \textbf{50.14} & \textbf{81.16} \\ \hline
\end{tabular}
}
\label{table2}
\end{table*}

\subsubsection{Impact of Different Pruning Ratios} 

In Table \ref{table3}, we present the zero-shot accuracy of the LLaMA3.1-8B model under different pruning configurations. The pruning removes 4 to 12 layers, corresponding to pruning ratios from 10.9\% to 32.6\%. The results reveal two main observations. First, the accuracy of all methods decreases as the pruning ratio increases. Second, our proposed methods, Ours (MSSD) and Ours (MASD), outperform existing baselines such as ShortGPT and EntroDrop across all settings. Under mild pruning with 4 layers removed, Ours (MSSD) achieves an average accuracy of 56.62\%, retaining 90.20\% of the original performance. When removing 8 and 10 layers, ShortGPT shows a drop in accuracy to 36.81\% and 39.21\%. In contrast, Ours (MSSD) achieves 49.01\% and 43.36\% under the same pruning ratios. Under the most aggressive pruning with 12 layers removed, Ours (MASD) remains more stable, reaching 41.12\% accuracy and retaining 65.65\% of the performance. These results indicate that our methods perform well across different pruning levels. We observe that Ours (MSSD) is better suited for moderate pruning, while Ours (MASD) is more robust under aggressive compression. Overall, our methods retain over 60\% of task performance without any healing, significantly outperforming existing structured pruning techniques.

\begin{table*}[t]
\centering
\caption{Comparison of zero-shot accuracy of LLaMA3.1-8B under different pruning settings. AVG is the average performance of the model on the zero-shot datasets. Retained performance (RP) represents the percentage of the original model’s performance retained by the pruning method. The best performance result is indicated in bold.}
\resizebox{\textwidth}{!}{
\begin{tabular}{c|ccccccccccc}
\hline
\small{Layer/Ratio} & \small{Method} & \small{BoolQ} & \small{PIQA} & \small{HeSW} & \small{WinoG} & \small{ARC-e} & \small{ARC-c} & \small{OBQA} & \small{MTQA} & \small{AVG} & \small{RP} \\ \hline
0/0\% & Dense & 82.11 & 80.25 & 60.01 & 73.88 & 81.48 & 51.45 & 33.40 & 39.60 & 62.77 & 100.0 \\ \hline
\multirow{4}{*}{4/10.9\%} & ShortGPT & 70.55 & 75.73 & 55.04 & 70.32 & 72.22 & 44.2 & 31.20 & 32.60 & 56.48 & 89.98 \\
 & EntroDrop & 62.20 & 72.58 & 40.91 & 64.64 & 61.28 & 38.82 & 29.40 & 33.07 & 50.36 & 80.23 \\
 & Ours (MSSD) & 69.79 & 74.65 & 53.54 & 72.22 & 74.96 & 43.86 & 30.20 & 33.74 & \textbf{56.62} & \textbf{90.20} \\
 & Ours (MASD) & 70.76 & 74.92 & 53.73 & 71.43 & 75.08 & 42.66 & 28.60 & 32.33 & 56.19 & 89.51 \\ \hline
\multirow{4}{*}{6/16.3\%} & ShortGPT & 71.99 & 72.52 & 49.98 & 68.51 & 63.93 & 39.59 & 25.40 & 30.08 & 52.75 & 84.04 \\
 & EntroDrop & 61.35 & 70.57 & 41.33 & 65.82 & 57.58 & 35.84 & 29.40 & 26.63 & 48.56 & 77.36 \\
 & Ours (MSSD) & 70.37 & 71.82 & 49.4 & 70.32 & 66.96 & 39.51 & 26.60 & 29.15 & \textbf{53.02} & \textbf{84.47} \\
 & Ours (MASD) & 69.33 & 72.03 & 49.18 & 69.30 & 67.34 & 38.99 & 26.40 & 29.38 & 52.74 & 84.02 \\ \hline
\multirow{4}{*}{8/21.7\%} & ShortGPT & 37.55 & 60.50 & 28.21 & 54.3 & 42.85 & 27.30 & 18.00 & 25.76 & 36.81 & 58.64 \\
 & EntroDrop & 62.11 & 68.77 & 39.88 & 65.98 & 53.70 & 33.79 & 24.00 & 25.39 & 46.70 & 74.40 \\
 & Ours (MSSD) & 71.80 & 67.85 & 43.84 & 68.98 & 56.57 & 35.32 & 21.20 & 26.50 & \textbf{49.01} & \textbf{78.08} \\
 & Ours (MASD) & 71.80 & 67.85 & 43.84 & 68.98 & 56.57 & 35.32 & 21.20 & 26.50 & \textbf{49.01} & \textbf{78.08} \\ \hline
\multirow{4}{*}{10/27.2\%} & ShortGPT & 52.17 & 60.94 & 27.64 & 57.62 & 44.15 & 28.75 & 18.2 & 24.22 & 39.21 & 62.47 \\
 & EntroDrop & 62.23 & 65.67 & 38.17 & 62.27 & 46.21 & 26.71 & 23.00 & 22.24 & 43.31 & 69.00 \\
 & Ours (MSSD) & 64.77 & 63.87 & 36.93 & 59.83 & 48.36 & 33.19 & 21.00 & 18.96 & \textbf{43.36} & \textbf{69.08} \\
 & Ours (MASD) & 64.77 & 63.87 & 36.93 & 59.83 & 48.36 & 33.19 & 21.00 & 18.96 & \textbf{43.36} & \textbf{69.08} \\ \hline
\multirow{4}{*}{12/32.6\%} & ShortGPT & 55.96 & 60.12 & 30.62 & 55.09 & 40.03 & 27.56 & 17.80 & 22.28 & 38.68 & 61.62 \\
 & EntroDrop & 51.56 & 62.13 & 33.42 & 58.8 & 42.51 & 23.46 & 20.40 & 21.91 & 39.27 & 62.56 \\
 & Ours (MSSD) & 45.81 & 62.89 & 34.12 & 61.01 & 40.45 & 26.96 & 17.60 & 21.31 & 38.77 & 61.76 \\
 & Ours (MASD) & 63.09 & 62.35 & 33.71 & 56.04 & 42.85 & 31.06 & 21.40 & 19.20 & \textbf{41.21} & \textbf{65.65} \\ \hline
\end{tabular}
}

\label{table3}
\end{table*}

\subsubsection{Inference Speedup}

Our work reduces model's depth, which theoretically enhances runtime efficiency during inference. Since both Ours (MSSD) and Ours (MASD) remove the same number of layers, they are expected to yield identical theoretical speedups. To validate the practical acceleration benefits, we perform inference evaluation on the WikiText2 dataset, comparing the model pruned with Ours (MSSD) against the dense model. Each input sequence contains 4 tokens, and the model generates 2048 tokens per instance with a batch size of 16. The evaluation is repeated 10 times on a single A40 GPU, and the final throughput is reported as the average across all runs. As shown in Table \ref{throughput}, we present the inference throughput and speedup results for the LLaMA3.1-8B model under various layer pruning settings. The results exhibit a clear positive correlation between the extent of pruning and inference efficiency. The dense model achieves a throughput of 97.40 tokens per second. As more layers are pruned, both throughput and speedup increase consistently. For example, pruning 6 layers leads to a 1.17× speedup, while pruning 12 layers significantly boosts the throughput to 144.79 tokens per second, corresponding to a 1.49× speedup. These findings demonstrate that our depth pruning method can effectively and predictably accelerate inference by directly reducing the computational cost of the forward pass. This capability is particularly valuable for real-world deployment of large language models, enabling lower latency and improved service efficiency.

\begin{table}[t!]
\centering
\caption{Throughput of LLaMA3.1-8B under different pruning ratios on a single A40 GPU.}
\begin{tabular}{cccc}
\hline
Layer & Ratio & Throughput & \multicolumn{1}{l}{Speedup} \\ \hline
0 & 0.00\% & 97.40 & 1.00x \\
4 & 10.9\% & 103.22 & 1.05x \\
6 & 16.3\% & 114.41 & 1.17x \\
8 & 21.7\% & 125.46 & 1.29x \\
10 & 27.2\% & 132.41 & 1.36x \\
12 & 32.6\% & 144.79 & 1.49x \\ \hline
\end{tabular}
\label{throughput}
\end{table}

\subsubsection{Fine-Tuning Performance}

\begin{figure}[t!]
\centering
\includegraphics[width=0.8\columnwidth]{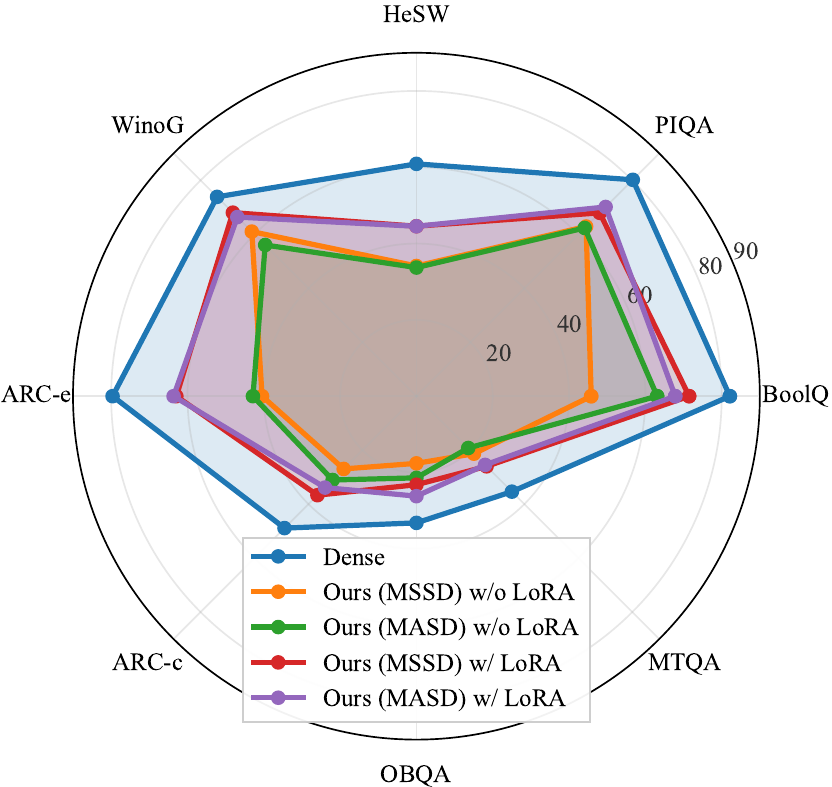} 
\caption{LoRA fine-tuning performance of the LLaMA3.1-8B model pruned by 12 layers.}
\label{finetuning}
\end{figure}

In Figure \ref{finetuning}, we evaluate our method by removing 12 layers (a 32.6\% compression rate) from LLaMA3.1-8B. The baseline Dense model achieves an average performance of 61.78. In contrast, Ours (MSSD)  performance drops to 38.77\% (61.76\% RP), while Ours (MASD) retains a higher score of 41.21\% (66.70\% RP). Notably, the crucial finding is the model's remarkable capacity for recovery. After applying just a single epoch of lightweight Low-Rank Adaptation (LoRA) fine-tuning, the performance of both Ours (MSSD) and Ours (MASD) is substantially restored, with their RP surging from approximately 65\% to over 80\% (81.10\% and 80.54\%, respectively). A deeper analysis of individual benchmarks reveals that this recovery is particularly effective on complex reasoning tasks. Notably, on datasets like BoolQ, PIQA, and WinoG, the performance of the healed model closely approaches that of the original dense model. This targeted restoration of capabilities, far beyond a simple uniform improvement, demonstrates a key advantage of SimDiff: it preserves the essential, learnable backbone of the model, ensuring the pruned architecture remains highly amenable to efficient post-training restoration of critical knowledge. 

\begin{figure}[!ht]
\centering
\includegraphics[width=0.8\columnwidth]{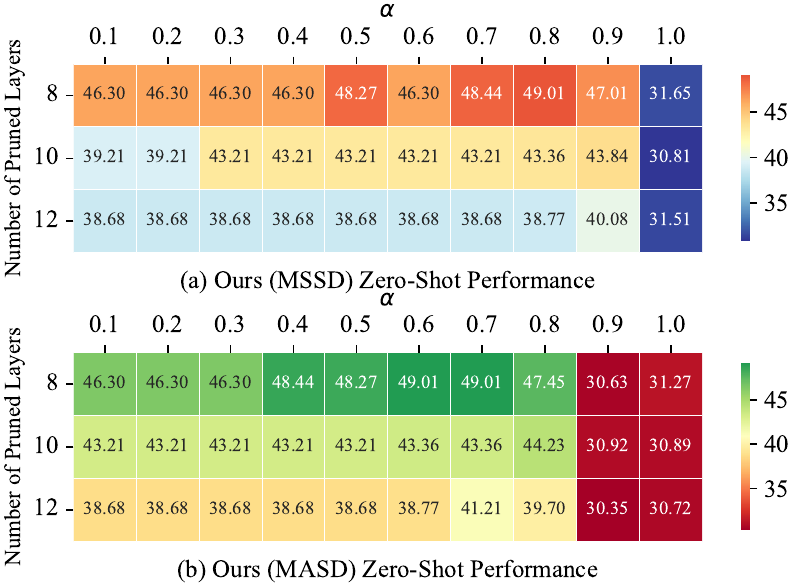} 
\caption{Pruning performance of LLaMA3.1-8B under different $\alpha$ values.}
\label{alpha}
\end{figure}


\subsubsection{Cross-Architecture Robustness}

\begin{figure*}[t]
\centering
\includegraphics[width=\columnwidth]{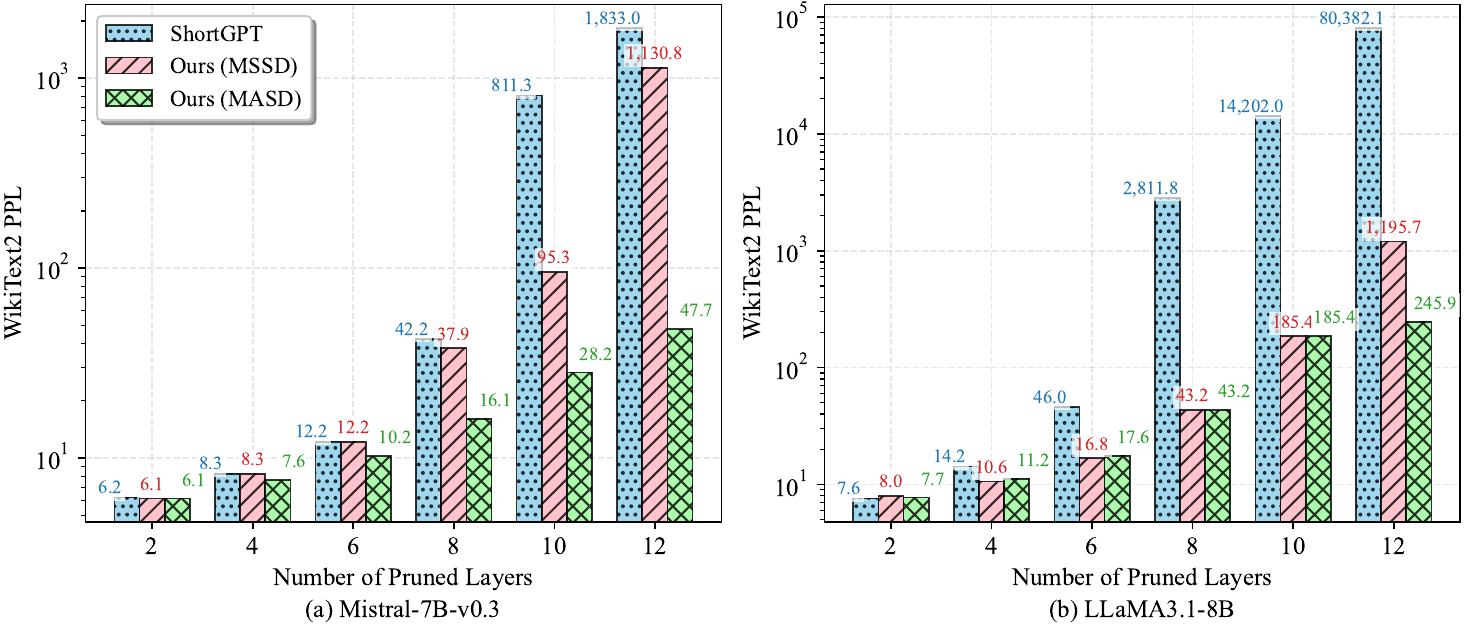} 
\caption{WikiText2 perplexity (PPL) for Mistral-7B-v0.3 and LLaMA3.1-8B across varying numbers of pruned layers. The y‑axis uses a logarithmic scale, and lower PPL indicates better language modeling performance.}
\label{valid}
\end{figure*}

As highlighted in Figure \ref{motivation} of Section \ref{sec:introduction}, the pruning performance of ShortGPT exhibits strong architectural dependence. To further validate and quantify this observation, Figure \ref{valid} presents a detailed comparison of language modeling performance, measured by WikiText2 perplexity, for Ours (MSSD) and Ours (MASD) against the ShortGPT baseline. The evaluation spans two models, Mistral-7B-v0.3 and LLaMA3.1-8B, under increasingly aggressive pruning depths from 2 to 12 layers. Due to the wide range of PPL values, the y-axis is on a logarithmic scale, where lower values indicate superior performance. Specifically, at low pruning ratios (e.g., removing up to 6 layers), the performance of all three methods is largely comparable on both models, with only minor differences in PPL.  On Mistral-7B-v0.3, ShortGPT’s PPL rises rapidly from 42.2 to 1,833.0 as 12 layers are pruned, while Ours (MSSD) and Ours (MASD) remain stable at 1,130.8 and 47.7, respectively. The gap is even more pronounced on LLaMA3.1-8B. These results confirm that SimDiff’s dual similarity–difference criterion effectively prevents catastrophic degradation and ensures stable performance under aggressive depth pruning.

\subsubsection{Performance on More Advanced LLMs}

To further validate the scalability and universality of our proposed pruning framework, we extend the experiments to a broader range of large language models, covering both compact and large-scale architectures, including Qwen2.5-0.5B, Qwen2.5-1.5B, Qwen2.5-3B, and LLaMA2-13B. These models differ significantly in size, training corpus, and architectural design, providing a rigorous test of cross-model robustness. On the smallest model, Qwen2.5-0.5B, aggressive pruning (up to 12 layers) severely degrades baseline methods such as ShortGPT, whose perplexity explodes to over 10,000, whereas Ours (MSSD) maintains a stable perplexity of 208.23 and achieves the highest average accuracy (33.89\%). For the medium-scale Qwen2.5-1.5B, our approach achieves consistent gains of 2–3\% absolute accuracy over ShortGPT across all pruning ratios, with notably lower perplexities (e.g., 26.34 vs.\ 64.45 under 10-layer pruning), confirming the efficiency and stability of our similarity-discrepancy criterion. On the larger Qwen2.5-3B configuration, both methods coincide under mild pruning but diverge under heavier compression, where Ours (MASD) sustains 41.42\% average accuracy, outperforming ShortGPT by 4.16\%. Finally, for the large scale LLaMA2-13B, our methods maintain the best overall results across all pruning levels: under 12-layer pruning, Ours (MASD) achieves 47.77\% average accuracy and the lowest perplexity (10.44), surpassing ShortGPT by 2.39\% while preserving high stability across all tasks. Collectively, these results confirm that the proposed SimDiff framework generalizes effectively across diverse architectures and parameter scales, consistently outperforming existing baselines in both accuracy and language modeling quality, even under aggressive compression. These findings also highlight a clear division between the two variants. MSSD performs better under moderate pruning by preserving layers with strong transformational impact, while MASD exhibits superior robustness under aggressive pruning due to its stability against outlier activations.

\begin{table*}[t]
\centering
\caption{Comparison of zero-shot accuracy of LLaMA3.1-8B under different pruning settings. AVG is the average performance of the model on the zero-shot datasets. When MSSD and MASD produce identical pruned layer selections, we uniformly refer to them as Ours. The best performance result is indicated in bold.}
\resizebox{\textwidth}{!}{
\begin{tabular}{ccccccccccccc}
\hline
\textbf{Model} & \textbf{Layers} & \textbf{Method} & \textbf{WikiText2}$\downarrow$ & \textbf{BoolQ} & \textbf{PIQA} & \textbf{HeSW} & \textbf{WinoG} & \textbf{ARC-e} & \textbf{ARC-c} & \textbf{OBQA} & \textbf{MTQA} & \textbf{AVG} \\
\multirow{7}{*}{Qwen2.5-0.5B} & \multirow{2}{*}8 & ShortGPT & 110.23 & 47.16 & 59.90 & 30.07 & 51.70 & 41.08 & 20.65 & 18.40 & 22.35 & 36.41 \\
 &  & Ours & \textbf{38.00} & 43.18 & 63.17 & 30.26 & 50.43 & 48.82 & 20.48 & 19.60 & 22.45 & \textbf{37.30} \\ \cline{2-13} 
 & \multirow{2}{*}{10} & ShortGPT & 858.52 & 42.97 & 58.27 & 28.28 & 49.80 & 36.95 & 21.25 & 16.00 & 21.94 & 34.43 \\
 &  & Ours & \textbf{87.70} & 41.19 & 60.94 & 28.97 & 51.38 & 44.61 & 18.52 & 17.00 & 22.58 & \textbf{35.65} \\ \cline{2-13} 
 & \multirow{3}{*}{12} & ShortGPT & 10916.14 & 37.80 & 55.55 & 26.25 & 49.17 & 29.88 & 21.67 & 17.60 & 19.56 & 32.19 \\
     &  & Ours (MSSD) & \textbf{208.23} & 38.47 & 57.78 & 28.26 & 50.28 & 39.94 & 20.22 & 15.40 & 20.77 & \textbf{33.89} \\
 &  & Ours (MASD) & 399.39 & 40.24 & 58.16 & 27.50 & 49.57 & 37.25 & 19.80 & 17.00 & 20.13 & 33.71 \\ \hline
\multirow{6}{*}{Qwen2.5-1.5B} & \multirow{2}{*}8 & ShortGPT & 37.70 & 61.65 & 64.91 & 35.62 & 50.59 & 53.28 & 27.99 & 22.80 & 26.20 & 42.88 \\
 &  & Ours & \textbf{18.42} & 62.23 & 70.13 & 37.36 & 53.59 & 62.16 & 26.88 & 25.40 & 24.62 & \textbf{45.30} \\ \cline{2-13} 
 & \multirow{2}{*}{10} & ShortGPT & 64.45 & 55.57 & 64.15 & 32.90 & 53.43 & 47.52 & 22.35 & 23.00 & 23.48 & 40.30 \\
 &  & Ours & \textbf{26.34} & 59.97 & 67.85 & 34.77 & 53.67 & 60.23 & 25.43 & 23.20 & 25.76 & \textbf{43.86} \\ \cline{2-13} 
 & \multirow{2}{*}{12} & ShortGPT & 634.94 & 46.61 & 59.25 & 29.34 & 50.28 & 36.24 & 23.29 & 18.20 & 20.80 & 35.50 \\
 &  & Ours & \textbf{44.34} & 39.17 & 64.04 & 30.89 & 52.25 & 48.74 & 22.10 & 18.20 & 22.71 & \textbf{37.26} \\ \hline
\multirow{6}{*}{Qwen2.5-3B} & \multirow{2}{*}8 & ShortGPT & 13.58 & 62.75 & 74.16 & 44.26 & 53.35 & 70.75 & 34.13 & 29.60 & 29.98 & 49.87 \\
 &  & Ours & 13.58 & 62.75 & 74.16 & 44.26 & 53.35 & 70.75 & 34.13 & 29.60 & 29.98 & 49.87 \\ \cline{2-13} 
 & \multirow{2}{*}{10} & ShortGPT & 26.34 & 59.97 & 67.85 & 34.77 & 53.67 & 60.23 & 25.43 & 23.20 & 25.76 & 43.86 \\
 &  & Ours & 26.34 & 59.97 & 67.85 & 34.77 & 53.67 & 60.23 & 25.43 & 23.20 & 25.76 & 43.86 \\ \cline{2-13} 
 & \multirow{2}{*}{12} & ShortGPT & 44.34 & 39.17 & 64.04 & 30.89 & 52.25 & 48.74 & 22.10 & 18.20 & 22.71 & 37.26 \\
 &  & Ours & \textbf{23.14} & 46.18 & 68.66 & 35.94 & 52.17 & 56.27 & 24.91 & 22.40 & 24.86 & \textbf{41.42} \\ \hline
\multirow{8}{*}{LLaMA2-13B} & \multirow{3}{*}{8} & ShortGPT & 8.30 & 62.17 & 73.88 & 50.99 & 65.43 & 61.24 & 36.35 & 26.60 & 26.16 & 50.35 \\
 &  & Ours (MSSD) & \textbf{7.43} & 62.23 & 74.21 & 52.23 & 67.72 & 62.88 & 38.31 & 27.20 & 26.60 & \textbf{51.42} \\
 &  & Ours (MASD) & 7.45 & 62.42 & 74.92 & 52.44 & 66.77 & 60.73 & 37.80 & 27.80 & 27.24 & 51.26 \\ \cline{2-13} 
 & \multirow{2}{*}{10} & ShortGPT & 20.03 & 62.63 & 71.27 & 47.29 & 62.98 & 53.45 & 34.30 & 23.80 & 24.05 & 47.47 \\
 &  & Ours & \textbf{8.87} & 62.26 & 72.03 & 50.18 & 67.01 & 58.21 & 36.18 & 22.60 & 25.33 & \textbf{49.22} \\ \cline{2-13} 
 & \multirow{3}{*}{12} & ShortGPT & 34.71 & 62.39 & 68.44 & 45.49 & 62.27 & 48.02 & 31.91 & 20.20 & 24.32 & 45.38 \\
 &  & Ours (MSSD) & 11.50 & 62.45 & 69.86 & 46.75 & 65.19 & 52.95 & 32.17 & 21.00 & 24.82 & 46.90 \\
 &  & Ours (MASD) & \textbf{10.44} & 62.45 & 71.82 & 47.30 & 66.61 & 54.88 & 33.96 & 20.80 & 24.32 & \textbf{47.77} \\ \hline
\end{tabular}
}
\end{table*}

\subsubsection{Impact of Hyperparameter}

To investigate the optimal balance between our proposed similarity and difference metrics, we conduct an ablation study. As shown in Figure \ref{alpha}, we evaluate the zero-shot average performance of Ours (MSSD) and Ours (MASD) at different pruning depths (8, 10, and 12 layers) across various values of the hyperparameter $\alpha$ from 0.1 to 1. By our method's definition, a larger $\alpha$ value corresponds to a higher weight assigned to our proposed difference metric in the final importance score. For Ours (MSSD), performance generally improves or remains stable as $\alpha$ increases, reaching its optimum at $\alpha$ values near 0.9. This indicates that when using MSSD, placing a greater emphasis on our proposed difference metric is consistently beneficial, as it effectively captures layers that perform critical, high-magnitude transformations. In contrast, Ours (MASD)  exhibits a different pattern: its performance peaks in the mid-to-high $\alpha$ range of 0.6 to 0.8, but suffers a catastrophic collapse when $\alpha$ is 0.9. This suggests that while the difference is an important signal, relying on it almost exclusively is detrimental. These findings validate our motivation for fusing two orthogonal metrics. This highlights that neither the similarity nor the difference metric alone is sufficient to create a universally robust pruning criterion.

\begin{table}[t]
\centering
\caption{Post-training parameters}
\begin{tabular}{cc}
\hline
Parameter & Value \\ \hline
Dataset & Alpaca \\
Batch Size & 32 \\
Mirco Batch Size & 4 \\
Epoch & 1 \\
Learning Rate & 0.003 \\
LoRA-Rank & 8 \\
LoRA-Alpha & 16 \\
LoRA-Dropout & 0.05 \\
Max Length & 256 \\
Validation Dataset Size & 2000 \\
LoRA Target Modules & \begin{tabular}[c]{@{}c@{}}q\_proj, v\_proj, k\_proj, o\_proj, \\ down\_proj, up\_proj, gate\_proj\end{tabular} \\
Train on Inputs & True \\
Add EOS Token & False \\ \hline
\end{tabular}

\label{lora}
\end{table}

\section{Conclusion}

We proposed SimDiff, a unified depth pruning framework that evaluates layer importance from two complementary perspectives: representational similarity and transformation difference. The two metrics, MSSD and MASD, respectively capture outlier-sensitive and stable transformation behaviors, providing a balanced and interpretable criterion for identifying redundant layers. An adaptive weighting mechanism, optimized via ternary search, enables SimDiff to generalize effectively across architectures and parameter scales ranging from 0.5B to 13B parameters. Experimental results on LLaMA2, LLaMA3.1, Mistral, and Qwen2.5 models demonstrate that SimDiff consistently surpasses existing pruning baselines, achieving substantial inference acceleration and efficient recovery with minimal fine-tuning. Overall, SimDiff offers a robust and scalable solution for efficient LLM compression and deployment.

\section*{Acknowledgements}

We would like to thank the anonymous reviewers for their thoughtful comments and support on this work. This work was supported in part by the National Key Research and Development Program of China under Grant 2022YFF0902701; in part by the National Natural Science Foundation of China under Grants U21A20468, 62372058, U22A2026.

\appendix

\section{Post-Training Details}
\label{section:lora}

 To accelerate the recovery process and improve efficiency under limited data conditions, we adopt LoRA for fine-tuning the pruned models. The models are trained for one epoch on the Alpaca dataset with a batch size of 32. We use AdamW as the optimizer and set the learning rate to $3 * 10^{-4}$. The total runtime for LoRA fine-tuning is approximately 142 minutes. Table \ref{lora} provides the detailed conﬁgurations of post-training compensation. Table \ref{lora_tab} quantifies the performance recovery of the LLaMA3.1-8B model via lightweight LoRA fine-tuning after an aggressive pruning of 12 layers. A particularly noteworthy phenomenon is the performance reversal: while Ours (MASD) is superior immediately after pruning, Ours (MSSD) achieves the highest final score after LoRA recovery (AVG 50.10\% vs. 49.76\%). The results suggest that the MSSD criterion, by preserving layers with high-magnitude transformations, may yield a structural backbone that, while initially more fragile, is more amenable to fine-tuning and knowledge restoration. In summary, this experiment demonstrates that our pruning method preserves the model's core structural integrity, allowing for the efficient recovery of the vast majority of its performance with a lightweight healing process.

\begin{table*}[t]
\centering
\caption{LoRA fine-tuning performance of the LLaMA3.1-8B model pruned by 12 layers. AVG is the average performance of the model on the zero-shot datasets. Retained performance (RP) represents the percentage of the original model’s performance retained by the pruning method.}
\resizebox{\textwidth}{!}{
\begin{tabular}{ccccccccccc}
\hline
\textbf{Method} & \textbf{BoolQ} & \textbf{PIQA} & \textbf{HeSW} & \textbf{WinoG} & \textbf{ARC-e} & \textbf{ARC-c} & \textbf{OBQA} & \textbf{MTQA} & \textbf{AVG} & \textbf{RP} \\ \hline
Dense & 82.14 & 80.20 & 60.89 & 73.88 & 79.63 & 48.89 & 33.20 & 35.38 & 61.78 & 100.0 \\
Ours (MSSD) w/o LoRA & 45.81 & 62.89 & 34.12 & 61.01 & 40.45 & 26.96 & 17.60 & 21.31 & 38.77 & 61.76 \\
Ours (MASD) w/o LoRA & 63.09 & 62.35 & 33.71 & 56.04 & 42.85 & 31.06 & 21.40 & 19.20 & 41.21 & 66.70 \\
Ours (MSSD) w/ LoRA & 71.47 & 67.95 & 44.55 & 68.03 & 62.92 & 36.69 & 23.20 & 26.00 & 50.10 & 81.10 \\
Ours (MASD) w/ LoRA & 67.83 & 70.13 & 44.53 & 66.38 & 63.64 & 33.87 & 26.20 & 25.46 & 49.76 & 80.54 \\ \hline
\end{tabular}
}
\label{lora_tab}
\end{table*}

\bibliographystyle{elsarticle-num} 
\bibliography{ref}
\end{document}